\documentclass[5p,twocolumn]{elsarticle}
\usepackage{graphicx}
\usepackage{amsmath,amssymb} % define this before the line numbering.
\usepackage{color}
\usepackage{amsmath,mathtools}
\usepackage[table]{xcolor}
\usepackage{multirow}
\usepackage{soul}
\usepackage{subcaption}
\usepackage{tabularx}
\usepackage[numbers]{natbib} 
\usepackage{epstopdf}
\usepackage{lineno}
\usepackage[hidelinks]{hyperref}
\usepackage{arydshln} % dashline

\makeatletter
\def\ps@pprintTitle{%
 \let\@oddhead\@empty
 \let\@evenhead\@empty
 \def\@oddfoot{}%
 \let\@evenfoot\@oddfoot}
\makeatother

\newcommand{\WKS}[1]{\textcolor{black}{#1}}
\newcommand{\liong}[1]{\textcolor{black}{#1}}
\newcommand{\christy}[1]{\textcolor{black}{#1}}
\newcommand{\christyy}[1]{\textcolor{black}{#1}}
\newcommand{\ct}[1]{\textcolor{black}{#1}}
\newcommand{\john}[1]{\textcolor{black}{#1}}
\newcommand{\ReviewerA}[1]{\textcolor{black}{#1}}
\newcommand{\ReviewerB}[1]{\textcolor{black}{#1}}
\newcommand{\ReviewerC}[1]{\textcolor{black}{#1}}
\newcommand{\modified}[1]{\textcolor{black}{#1}}
\newcommand{\modifiedA}[1]{\textcolor{black}{#1}}
\newcommand{\modifiedB}[1]{\textcolor{black}{#1}}

\journal{Signal Processing: Image Communication}

\biboptions{sort&compress}

\begin{document}  

\begin{frontmatter}
\title{Less is More: Micro-expression Recognition from Video using Apex Frame}

  \author[add1]{Sze-Teng Liong} %[add1, add3]
  \ead{christyliong91@gmail.com}
  \author[add2]{John~See}
  \ead{johnsee@mmu.edu.my}
  \author[add3]{KokSheik Wong\corref{cor1}}
  \ead{wong.koksheik@monash.edu}
  \author[add4]{Raphael C.-W. Phan}
  \ead{raphael@mmu.edu.my}

  \cortext[cor1]{Corresponding author}
  \address[add1]{Department of Electronic Engineering, Feng Chia University, Taichung 40724, Taiwan R.O.C.}
%  \address[add1]{Faculty of Information and Communication Technology, Xiamen University Malaysia, 43900 Sepang, Selangor, Malaysia} 	
  \address[add2]{Faculty of Computing and Informatics, Multimedia University, 63100 Cyberjaya, Malaysia} 	
  \address[add3]{School of Information Technology, Monash University Malaysia, 47500 Selangor, Malaysia}
  \address[add4]{Faculty of Engineering, Multimedia University, 63100 Cyberjaya, Malaysia}

\begin{abstract}
Despite recent interest and advances in facial micro-expression research, there is still plenty of room for improvement in terms of micro-expression recognition. Conventional feature extraction approaches for micro-expression video consider either the whole video sequence or a part of it, for representation. However, with the high-speed video capture of micro-expressions (100-200 fps), are all frames necessary to provide a sufficiently meaningful representation? Is the luxury of data a bane to accurate recognition? A novel proposition is presented in this paper, whereby we utilize only two images per video, namely, the apex frame and the onset frame. The apex frame of a video contains the highest intensity of expression changes among all frames, while the onset is the perfect choice of a reference frame with neutral expression. A new feature extractor, Bi-Weighted Oriented Optical Flow (Bi-WOOF) is proposed to encode essential expressiveness of the apex frame. We evaluated the proposed method on \modified{five} micro-expression databases--- \modified{CAS(ME$)^2$}, CASME II, SMIC-HS, SMIC-NIR and SMIC-VIS. Our experiments lend credence to our hypothesis, with our proposed technique achieving a state-of-the-art F1-score recognition performance of 0.61 and 0.62 in the high frame rate CASME II and SMIC-HS databases respectively.
\end{abstract}


\end{frontmatter}

%\linenumbers

% sections in separate files
\section{Introduction} 
Have you ever thought that someone was lying to you, but have no evidence to prove it? Or have you always found it difficult to interpret one's emotion? Recognizing micro-expressions could help to solve these doubts. 

\emph{Micro-expression} is a very brief and rapid facial emotion that is provoked involuntarily \cite{ekman1969nonverbal}, revealing a person's true feelings. Akin to normal facial expression, also known as \emph{macro-expression}, it can be categorized into six basic emotions: happy, fear, sad, surprise, anger and disgust. However, macro-expressions are easily identified in real-time situations with the naked eye as it occurs between 2--3 seconds and can be found over the entire face region. On the other hand, a micro-expression is both \emph{micro} (short duration) and \emph{subtle} (small intensity) \cite{ekman1971constants} in nature. It lasts between $1/5$ to $1/25$ of a second and usually occurs in only a few parts of the face. These are the main reasons why people are sometimes unable to realize or recognize the genuine emotion shown on a person's face \cite{ekman2009lie,porter2008reading}. Hence, the ability to recognize micro-expressions is beneficial in both our mundane lives and also society at large. At a personal level, we can differentiate if someone is telling the truth or lie. Also, analyzing a person's emotions can help facilitate understanding of our social relationships, while we are increasingly awareness of the emotional states of our own selfs and of the people around us. More essentially, recognizing these micro-expressions is useful in a wide range of applications, including psychological and clinical diagnosis, police interrogation and national security \cite{frank2009see, sullivan2009police, frank2009protect}. 

\begin{figure}[t!]
\centering
\includegraphics[width=80mm]{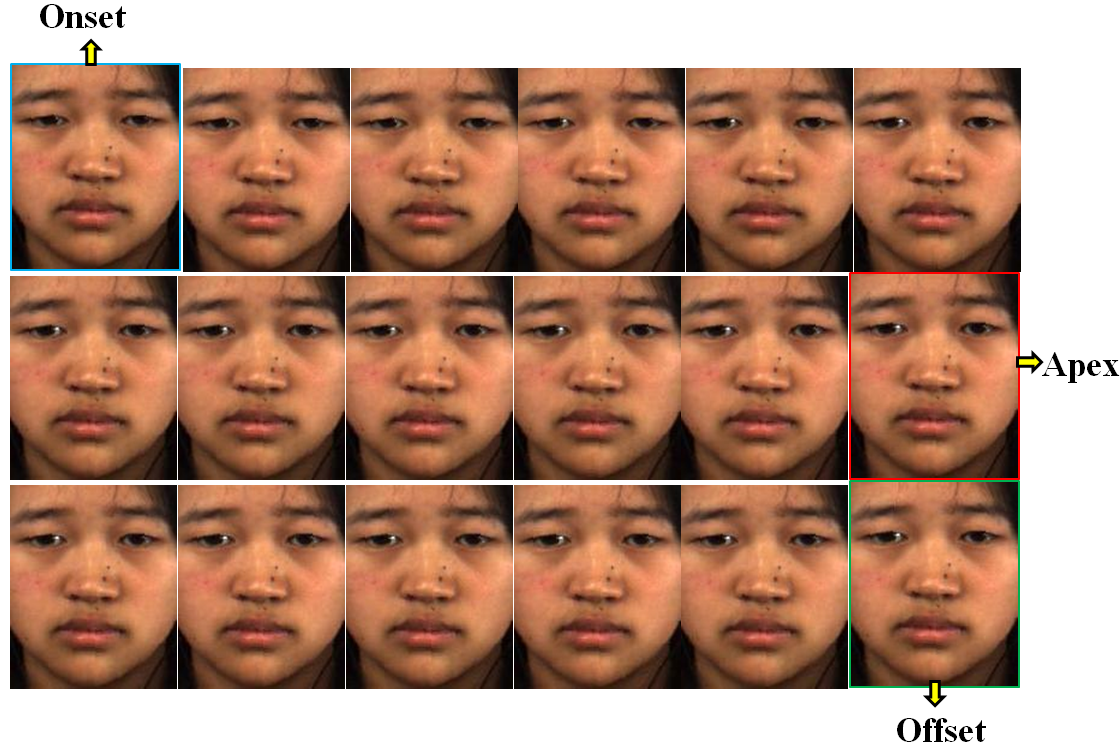}
\caption{Example of a sequence of image frames (ordered from left to right, top to bottom) of a \christy{surprise} expression from the CASME II \protect\cite{casme2} database, with the onset, apex and offset frame indications}% \WKS{Looks pretty sad to me. Please check if this is the correct frame sequence.}\christy{ Thank you for pointing out this. It should be surprise.(CASME II, sub02, EP11 01)}} 
\label{fig:example}
\end{figure}

Micro-expression was first discovered by psychologists, Ekman and Friesen \cite{ekman1969nonverbal} in 1969, from a case where a patient was trying to conceal his sad feeling by covering up with smile. They detected the patient's genuine feeling by carefully observing the subtle movements on his face, and found out that the patient was actually planning to commit suicide. Later on, they established Facial Action Coding System (FACS) \cite{ekman1978facs} to determine the relationship between facial muscle changes and emotional states. This system can be used to identify the exact time each action unit (AU) begins and ends. The occurrence of the first visible AU is called the onset, while that of the disappearance of the AU is the offset. Apex is the point when the AU reaches the peak or the highest intensity of the facial motion. The timings of the onset, offset and apex for the AUs may differ for the same emotion type. Figure \ref{fig:example} shows a sample sequence containing frames of a \christy{surprise} expression from a micro-expression database, with the indication of onset, apex and offset frames.

\section{Background}

Micro-expression analysis is arguably one of the lesser explored areas of research in the field of machine vision and computational intelligence. Currently, there are less than fifty micro-expressions related research papers published since 2009.
While databases for normal facial expressions are widely available \cite{anitha2010survey}, facial micro-expression data, particularly those of spontaneous nature, is somewhat limited for a number of reasons. Firstly, the elicitation process demands for good choice of emotional stimuli that has high ecological validity. Post-capture, the labeling of these micro-expression samples require the verification of psychologists or trained experts. Early attempts centered on the collection of posed micro-expression samples, i.e. USF-HD \cite{shreve2009towards} and Polikovsky's \cite{polikovsky} databases, which went against the involuntary and spontaneous nature of micro-expressions \cite{ekman2007emotions}. 
%On the contrary, feature extraction methods and databases for macro-expression studies are well-developed and established. 
%dfdf
Thus, the lack of spontaneous micro-expression databases had hindered the progress of micro-expression research. 
%Moreover, the subtlety of a micro-expression's intensity and the quickness of its occurrence, are also natural obstacles that can prevent proper elicitation and labeling of data samples. 
%
Nonetheless, since 2013, the emergence of three prominent spontaneous facial micro-expression databases -- the SMIC from University of Oulu \cite{smic} and the CASME/ CASME II/ \modified{CAS(ME$)^2$} \cite{yan2013casme, casme2, qu2017casme} from the Chinese Academy of Sciences, have breathed fresh interest into this domain.

\begin{figure}[t!]
\centering
\includegraphics[width=\linewidth]{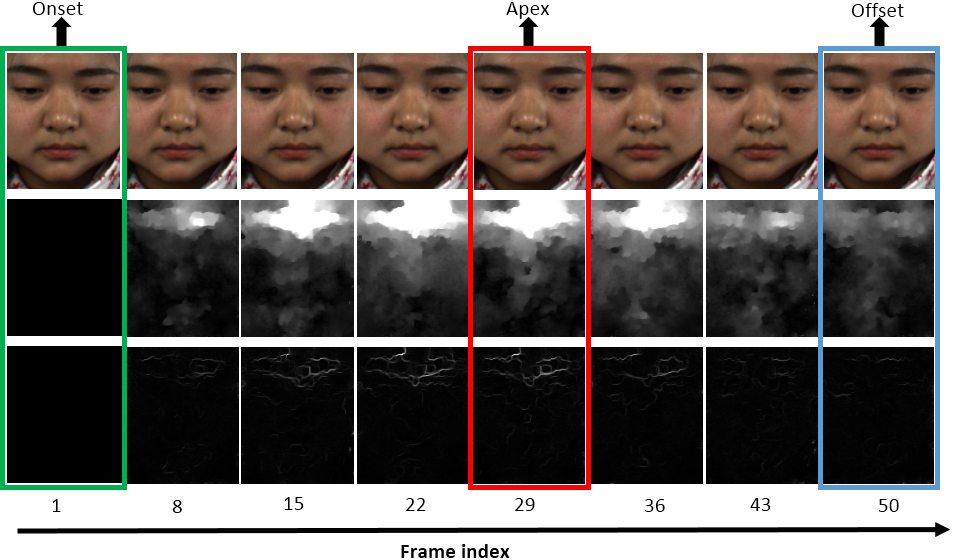}
\caption{\ReviewerB{Illustration of (top row) original images; (middle row) optical flow magnitude computed between the onset and subsequent frames; and (bottom row) optical strain computed between the onset and subsequent frames
\label{fig:of_os}}}%sub05_EP04_05
\end{figure}

\ReviewerB{There are two primary tasks in an automated micro-expression system, i.e., spotting and recognition. The former identifies a micro-expression occurrence (and its interval of occurrence), or to locate some important frame instances such as onset, apex and offset frames (see Figure \ref{fig:example}). Meanwhile, the latter classifies the expression type given the ``spotted" micro-expression video sequence. A majority of works focused solely on the recognition task of the system, whereby new feature extraction methods have been developed to improve on micro-expression recognition rate. 
Figure~\ref{fig:of_os} illustrates the optical flow magnitude and optical strain magnitude computed between the onset (assumed as neutral expression) and subsequent frames.
It is observed that the apex frames (middle and bottom rows in Figure~\ref{fig:of_os}) are the frames with the highest motion changes (bright region) among the video sequence.} 

Micro-expression databases are pre-processed before releasing to the public. This process includes face registration, face alignment and ground-truth labeling (i.e., AU, emotion type, frame indices of onset, apex and offset). In the two most popular spontaneous micro-expression databases, namely the CASME II \cite{casme2} and SMIC \cite{smic}, the first two processes (face registration and alignment) were achieved automatically. Active Shape Model (ASM) \cite{cootes1995active} is used to detect a set of facial landmark coordinates; then the faces are transformed based on the template face according to its landmark points using the classic Local Weighted Mean (LWM) \cite{gosh1988image} method. %Note that both ASM and LWM are fully automated. 
However, the last process, i.e., ground-truth labeling, is not automatic and requires the help of psychologists or trained experts. In other words, the annotated ground-truth labels may vary depending on the coders. As such, the reliability and consistency of the markings are less than ideal, which may affect the recognition accuracy of the system. 

\subsection{Micro-expression recognition}

\ReviewerA{Recognition baselines for the SMIC, CASME II \modified{and CAS(ME$)^2$} databases were established with the original works \cite{smic,casme2,qu2017casme}  with Local Binary Patterns-Three Orthogonal Planes (LBP-TOP)\liong{~\cite{dynamic}} as the choice of spatio-temporal descriptor, and Support Vector Machines (SVM)\liong{~\cite{suykens1999least}} as classifier. Subsequently, a number of LBP variants\liong{~\cite{wang2015lbp,huang2015facial,huang2016spontaneous}} were proposed to improve on the usage of LBP-TOP. Wang et al. \cite{wang2015lbp} presented an efficient representation that reduces the inherent redundancies within LBP-TOP, while Huang et al. \cite{huang2015facial} adopted an integral projection method to boost the capability of LBP-TOP by supplementing shape information. More recently, another LBP variant called SpatioTemporal Completed Local Quantization Pattern (STCLQP) \cite{huang2016spontaneous} was proposed to extract three kinds of information (local sign, magnitude, orientation) before encoding them into a compact codebook. A few works stayed away from using conventional pixel intensity information in favor of other base features such as optical strain information \cite{liong2014optical,liong2014subtle} and monogenic signal components \cite{oh2015monogenic}, before describing them with LBP-TOP. \modifiedB{There were other methods proposed that derived useful features directly from color spaces \cite{wang2015micro} and optical flow orientations \cite{liu2016main}.}}

\modifiedB{Two most recent works~\cite{le2016sparsity,xu2016microexpression} presented alternative schemes to deal with the minute changes in micro-expression videos Le et al. \cite{le2016sparsity} hypothesized that dynamics in subtle occurring expressions contain a significantly large number of redundant frames, therefore they are likely to be ``sparse". 
Their approach determines the optimal vector of amplitudes with a fixed sparsity structure and recognition performance reportedly significantly better than using the standard Temporal Interpolation Model (TIM) \cite{tim}.
Xu et al. \cite{xu2016microexpression} characterized the local movements of a micro-expression by the principal optical flow direction of spatiotemporal cuboids extracted at a chosen granularity.  
On the other hand, the works by~\cite{ben2016gait, zhang2016micro,wang2014micro} reduce the dimensionality of the features extracted from micro-expression videos using Principal Component Analysis (PCA), while \cite{wang2016main} employed sparse tensor analysis to minimize the dimension of features.} 

\subsection{Micro-expression spotting}

There are several works which attempted to spot the temporal interval (i.e., onset-offset) containing micro-expressions from raw videos in the databases. By \emph{raw}, we refer to video clips in its original form, without any pre-processing. In \cite{moi2014spotting}, the authors searched for the frame indices that contain micro-expressions. They utilized Chi-Squared dissimilarity to calculate the distribution difference between the Local Binary Pattern (LBP) histogram of the current feature frame and the averaged feature frame. The frames which yield score greater than a predetermined threshold were regarded as frames with micro-expression. 

A similar approach was carried out by \cite{davison2015micro}, except that: (1) a denoising method was added before extracting the features, and; (2) the Histogram of Gradient was used instead of LBP. However, the database they tested on was not publicly available. Since the benchmark \WKS{video sequences} used in this paper \christy{\cite{davison2015micro}} %\WKS{This paper refers to which paper?} 
and that in \cite{moi2014spotting} are different, their performances cannot be compared directly. Both papers claimed that the eye blinking movement is one type of the micro-expression. However, it was not detailed in the ground-truth and hence the frames containing eye blinking movements were annotated manually. \modifiedB{A recent work by Wang et al.~\cite{wang2017main} proposed main directional maximal difference analysis for spotting facial movements from long-term videos.}

To the best of our knowledge, there is only one recent work that attempted to combine both spotting and recognition of micro-expressions, which is the work of Li et al. \cite{li2015reading}. They extended the work by Moilanen et al. \cite{moi2014spotting}, where after the spotting stage, the spotted micro-expression frames (i.e., those with the onset and offset information) were concatenated to a single sequence for expression recognition. In the recognition task, they employed motion magnification technique and proposed a new feature extractor - the Histograms of Image Gradient Orientation. However, the recognition performance was poor compared to the state-of-the-art. Besides, the frame rate of the database is 25~fps, which means that the maximum frame number in a raw video sequence is only \WKS{1/5 $s$ $\times$ 25~fps = $5$.}

\subsection{Apex spotting}

Apart from the aforementioned micro-expression frames searching approaches, the other technique used is to automatically spot the instance of the single apex frame in a video. The micro-expression information retrieved from that apex frame is expected to be insightful in both psychological and computer vision research purposes, because it contains the maximum facial muscle movements throughout the video sequence. Yan et al. \cite{yan2014quantify} published the first work in spotting the apex frame. They employed two feature extractors (i.e., LBP and Constraint Local Models) and reported the average frame distance between the spotted apex and the ground-truth apex. The frame that has the highest feature difference between the first frame and the subsequent frames is defined to be the apex. However, there are two flaws in this work: (1) The average frame distance calculated was not in absolute mean, which led to incorrect  results; (2) The method was validated by using only $\sim$ 20\% of the video samples in the database \christy{(i.e., CASME II)}%\WKS{how many databases are there?}
, hence not conclusive and convincing. 

The second work on apex frame spotting was presented by Liong et al. \cite{liong2015automatic}, which differs from the first work by Yan et al.~\cite{yan2014quantify} as follows: (1) A {\it divide-and-conquer} strategy was implemented to locate the frame index of the apex, because the maximum difference between the first and the subsequent frames might not necessarily be the apex frame; (2) An extra feature extractor was added to confirm the reliability of the method proposed; (3) Selected important facial regions were considered for feature encoding instead of the whole face, and; (4) All the video sequences in the database \christy{(i.e., CASME II) }%\WKS{how many databases are there?}
were used for evaluation and the average frame distance between the spotted and groundtruth apex were computed in absolute mean.

\modified{Later, Liong et al.~\cite{liong2016auto} spotted the micro-expression on {\it long videos} (i.e., SMIC-E-HS and CASME II-RAW databases). 
Specifically, {\it long video} refers to the raw video sequence which may include the frames with micro-expressions as well as irrelevant motion that are present before the onset and after the offset. 
On the other hand, {\it short video} is a sub-sequence of the {\it long video} starting from the onset and ending with the offset. 
In other words, all frames before the onset frame and after the offset frame are excluded.
A novel eye masking approach was also proposed to mitigate the issue where frames in the {\it long videos} may contain large and irrelevant movements such as eye blinking actions, which can potentially cause erroneous spotting.}

%The threshold technique was also used by Patel et al. \cite{patel2015spatio}. They first detect the peak frame of the video sequence using TV-L1 optical flow method and grouped the landmarks detected based on the AUs. Then, the onset and offset instances were detected using the peak information. The drawbacks of this work are: (1) The frames that have eye blinks are manually removed; (2) The grouping of the AUs are empirically decided; (3) Some frames in the video sequence are ignored.

\subsection{``Less" is More?}
\john{
Considering these developments, we pose the following intriguing question: With the high-speed video capture of micro-expressions (100-200 fps), are all frames necessary to provide a sufficiently meaningful representation? While the works of Li et al. \cite{smic} and Le Ngo et al. \cite{le2015subtle,le2016sparsity} showed that a reduced-size sequence can somewhat help retain the vital information necessary for a good representation, there are no existing investigations into the use of the apex frame. How meaningful is the so-called apex frame? Ekman \cite{ekman1993facial} asserted that a ``snapshot taken at an point when the expression is at its apex can easily convey the emotion message". A similar observation by Esposito \cite{esposito2007amount} earmarked the apex as ``the instant at which the indicators of emotion are most marked". Hence we can hypothesize that the apex frame offers the strongest signal that depicts the ``momentary configuration" \cite{ekman1993facial} of facial contraction.
}

In this paper, we propose a novel approach to micro-expression recognition, where for each video sequence, we encode features from the representative apex frame with the onset frame as the reference frame. The onset frame is assumed to be the neutral face and is provided in all micro-expression databases (e.g., \modified{CAS(ME$)^2$}, CASME II and SMIC) while the apex frame labels are only available in \modified{CAS(ME$)^2$} and CASME II. To solve the lack of apex information in SMIC, a binary search strategy was employed to spot the apex frame~\cite{liong2015automatic}. We renamed $binary$ $search$ to {\it divide-and-conquer} for a more general terminology to this scheme. Additionally, we introduce a new feature extractor called {\it Bi-Weighted Oriented Optical Flow} (Bi-WOOF), which is capable of representing the apex frame in a discriminative manner, emphasizing facial motion information at both bin and block levels. The histogram of optical flow orientations is weighted twice at different representation scales, namely, bins by the magnitudes of optical flow, and block regions by the magnitudes of optical strain. %Optical flow estimates the motion of objects between two images over time, which was found effective in a recently proposed micro-expression recognition system \cite{liu2016main}. Optical strain, which is an extension of the optical flow, provides more precise subtle and micro facial changes, as was proven useful in several works on micro-expression analysis \cite{shreve2009towards, shreve2015automatic, liong2014optical, liong2014subtle}.
We establish our proposition by proving empirically through a comprehensive evaluation that was carried out on four notable databases.

The rest of this paper is organized as follows. Section~\ref{sec:algorithm} explains the proposed algorithm in detail. The descriptions of the databases used are discussed in Section~\ref{sec:experiment}, followed by Section~\ref{sec:results} that reports the experiment results and discussion for the recognition of micro-expressions. Finally, conclusion is drawn in Section~\ref{sec:conclusion}.

%\section{Related Work}
%\label{sec:related}

%\section{Feature Extraction}

\section{Proposed Algorithm}
\label{sec:algorithm}
\begin{figure*}[t!]
\centering
\includegraphics[width=0.65\linewidth]{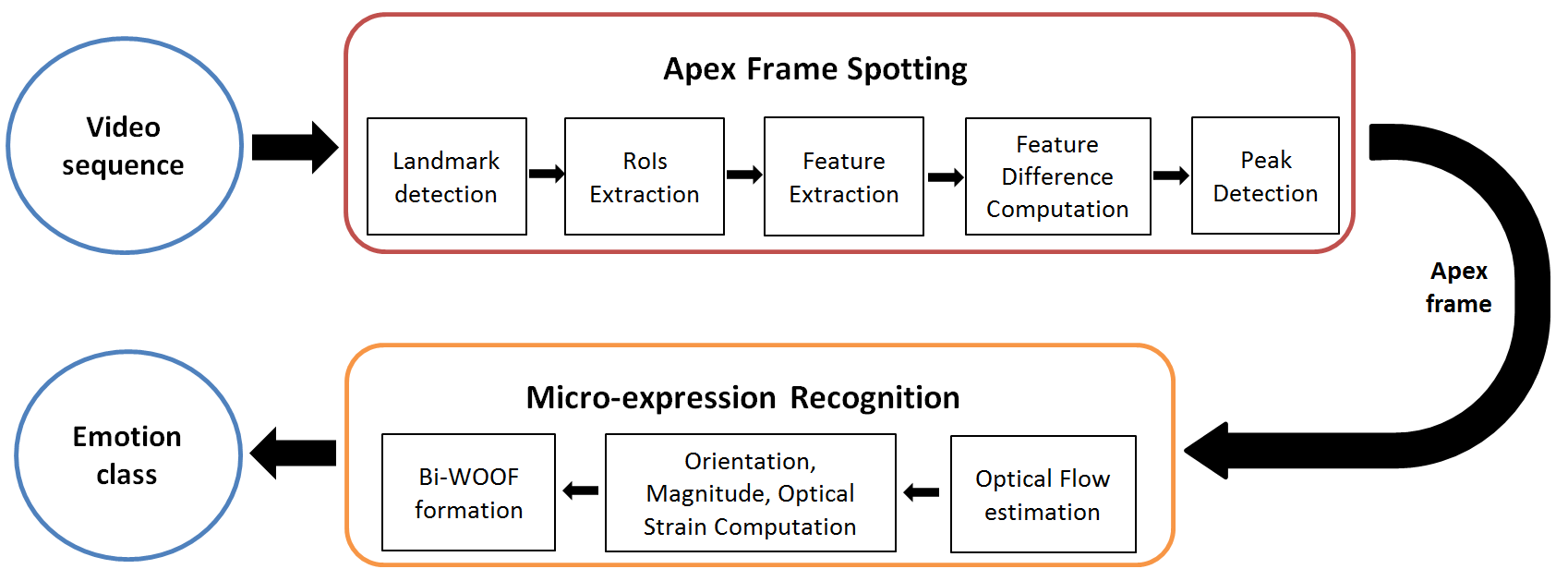}
\caption{Framework of the proposed micro-expression recognition system}
\label{fig:framework}
\end{figure*}

The proposed micro-expression recognition system comprises of two components, namely, apex frame spotting, and micro-expression recognition. The architecture overview of the system is illustrated in Figure~\ref{fig:framework}. The following subsections detail the steps involved.

\subsection{Apex Spotting}
\begin{figure*}[t!]
\centering
\includegraphics[width=0.8\linewidth]{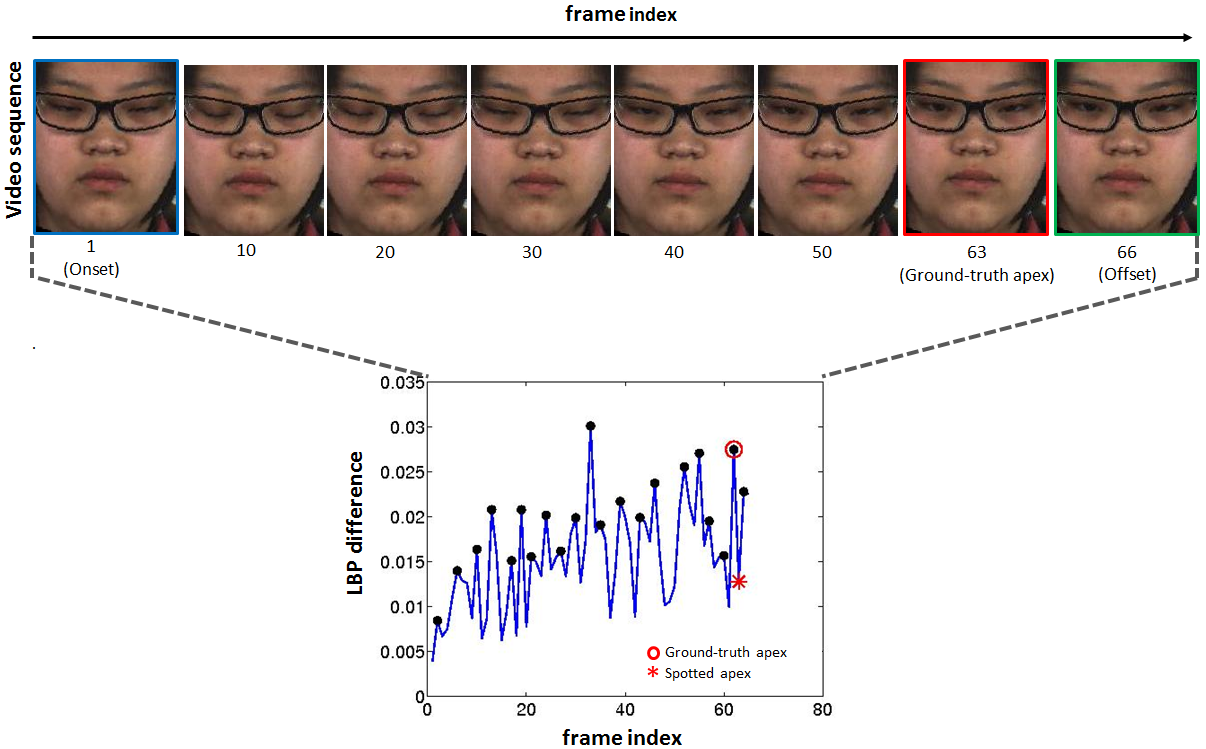}
\caption{\liong{Illustration of the apex spotting in the video sequence (i.e., sub20-EP12\_01 in CASME II~\protect\cite{casme2} database) using LBP feature extractor with {\it divide-and-conquer}~\protect\cite{liong2015automatic} strategy }}
\label{fig:example_spotting}
\end{figure*}

To spot the apex frame, we employ the approach proposed by Liong et al. \cite{liong2015automatic}, which consists of five steps: (1) The facial landmark points are first annotated by using a landmark detector called Discriminative Response Map Fitting (DRMF)~\cite{asthana2013robust}; (2) The regions of interest that indicate the facial region with important micro-expression details are extracted according to the landmark coordinates; (3) The LBP feature descriptor is utilized to obtain the features of each frame in the video sequence (i.e., from onset to offset); (4) The feature difference between the onset and the rest of the frames are computed using the correlation coefficient formula, and finally; (5) A peak detector with {\it divide-and-conquer} strategy is utilized to search for the apex frame based on the LBP feature difference.
\ReviewerC{Specifically, the procedures of {\it divide-and-conquer} methodology are: 
(A) The frame index of the peaks/ local maximum in the video sequence are detected by using a peak detector. 
(B) The frame sequence is divided into two equal halves (e.g., a 40 frames video sequence is split into two sub-sequences containing frame 1-20 and 21-40). 
(C) Magnitudes of the detected peaks are summed up for each of the sub-sequence. 
(D) The sub-sequence with the higher magnitude will be considered for the next computation step while the other sub-sequence will be discarded. 
(E) Steps (B) to (D) are repeated until the final peak (also known as apex frame) is found.}
Liong et al.~\cite{liong2015automatic} reported that the average estimated apex frame is 13 frames away from the ground-truths apex frames for {\it divide-and-conquer} methodology. 
Note that the micro-expression video has an average length of 68 frames.
Figure~\ref{fig:example_spotting} \liong{illustrates} the apex frame spotting approach in a sample video. It can be seen that, the ground-truth apex (frame \#63) and the spotted apex (frame \#64) differ only by one frame.

\subsection{Micro-expression Recognition}
\begin{figure*}[t!]
\centering
\includegraphics[width=0.6\linewidth]{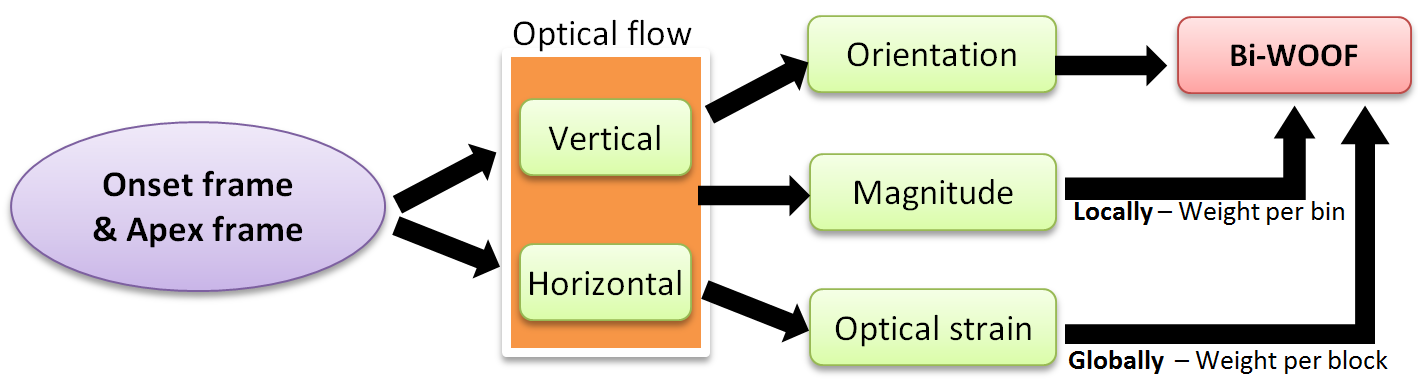}
\caption{Flow diagram of micro-expression recognition system}
\label{fig:flow}
\end{figure*}

Here, we discuss a new \christy{feature descriptor}%\WKS{ How about calling it 'feature descriptor?'}
, Bi-Weighted Oriented Optical Flow (Bi-WOOF) that represents a sequence of subtle expressions by using only two frames. As illustrated in Figure~\ref{fig:flow}, the recognition algorithm contains three main steps: (1) The horizontal and vertical optical flow vectors between the apex and neutral frames are estimated; (2) The orientation, magnitude and optical strain of each pixel's location are computed from the respective two optical flow components; (3) A Bi-WOOF histogram is formed based on the orientation, with magnitude locally weighted and optical strain globally weighted. 

\subsubsection{Optical flow estimation \cite{fleet2006optical}}%\WKS{need to cite a textbook here?}}
Optical flow approximates the changes of an object's position between two frames that are sampled at slightly different times. It encodes the motion of an object in vector notation, which indicates the direction and intensity of the flow of each image pixel. \ReviewerC{The horizontal and vertical components of the optical flow are defined as:} 

\ReviewerC{
\begin{equation} \label{eq:flow2}
    \vec{p} = [p = \frac{dx}{dt}, q = \frac{dy}{dt}]\textsuperscript{\it T},
\end{equation}}

\noindent\ReviewerC{
where ($dx$, $dy$) indicate the changes along the horizontal and vertical dimensions, and $dt$ is the change in time. %The principle of estimating the optical flow is, given a pixel in the first frame, look for the nearby pixel with the same color in the subsequent frame.
} 
%However, there are three assumptions when estimating the optical flow: (1) Brightness constancy: the observed appearance of an object in the image remains constant over time. Note that, we ignore the apparent motion that are caused by the lighting changes without any actual motion; (2) Spatial \WKS{coherency}: the neighboring pixels with similar pixel intensities in an image are likely to belong to the same region and move in a similar manner, and; (3) Temporal \WKS{persistency}: the motion of an object between two frames is small and hence the displacement is small.
\ReviewerC{The optical flow constraint equation is given by:
\begin{equation} \label{eq:flow}
\nabla I \bullet \vec{p} + I_{t} =0,
\end{equation}
\noindent where $\nabla I$ = {\it($I_{x},I_{y}$)} is the gradient vector of image intensity evaluated at $(x, y)$ and {\it $I_{t}$} is the temporal gradient of the intensity functions.}

We employed TV-L1 \cite{perez2013tvl1} for optical flow approximation due to its two major advantages, namely, better noise robustness and the ability to preserve flow discontinuities.
%This method is based on the minimization of a functional containing a data term using the L1 norm and a regularization term using the total variation of the flow. 

We first introduce and describe the notations which are used in the subsequent sections. A micro-expression video clip is denoted as: 
\begin{equation}
s_{i} = \{f_{i,j} | i=1,\dots,n; j=1,\dots ,F_{i}\},
\end{equation}
where $F_{i}$ is the total number of frames in the ${i}$-th sequence, which is taken from a collection of $n$ video sequences. For each video sequence, there is only one apex frame, $f_{i,a} \in f_{i,1},\dots ,f_{i,F{i}}$, and it can be located at any frame index.

The optical flow vectors of the onset (assumed as neutral expression) and the apex frames are predicted then denoted by $f_{i,1}$ and $f_{i,a}$ %\WKS{I modified the expression - still OK?} \christy{Ok}
, respectively.
Hence, each video of resolution $X\times Y$ produces only one set of optical flow map, expressed as:  
\begin{equation}
\nu_{i} = \{(u_{x,y},v_{x,y}) | x=1,\dots,X; y=1,\dots ,Y\}
\end{equation}
for $i \in 1,2, \dots n$.
\ReviewerC{Here, ($u_{x,y}$, $v_{x,y}$) are the displacement vectors in the horizontal and vertical directions respectively.}

\subsubsection{\ct{Computation of orientation, magnitude and optical strain}}
Given the optical flow vectors, we derive three %\WKS{\emph{characteristics} Why not the term `features'?}\christy{the word ``features'' might be confusing, as the main feature proposed is bi-woof}
 characteristics to describe the facial motion patterns: (1) magnitude: intensity of the pixel's movement; (2) orientation: direction of the flow motion, \ct{and}; (3) optical strain: subtle deformation intensity.

In order to obtain the magnitude and orientation, the flow vectors, $\vec{o}=(p,q)$%\WKS{I found that $\vec{P} = (p,q)$ is very confusing. Avoid using the same symbol on both left and right hand sides of the equations. How about $\vec{P} = (u,v)$, or simply $(a, b)$ or something along the line?} \christy{changed $\vec{p}$ to $\vec{o}$}
, are converted from euclidean coordinates to polar coordinates: 
\begin{equation}
\rho_{x,y} = \sqrt{ p{_{x,y}}^2 + q{_{x,y}}^2 },
\end{equation} and
\begin{equation}
\theta_{x,y} = tan^{-1} \frac{q_{x,y}}{p_{x,y}},
\end{equation}
where $\rho$ and $\theta$ are the magnitude and orientation, respectively.

The next step is to compute the optical strain, $\varepsilon$, based on the optical flow vectors.
For a \WKS{sufficiently} small facial pixel's movement, it is able to approximate the deformation intensity, also known as the infinitesimal strain tensor. In brief, the infinitesimal strain tensor is derived from the Lagrangian and Eulerian strain tensor after performing a geometric linearisation~\cite{simof2008computational} %\WKS{Please check for consistency - linearization (US) vs linearisation (UK)} \christy{done checking}
. In terms of displacements, the typical infinitesimal strain ($\varepsilon$) is defined as: 

\begin{equation} \label{eq:tensor}
\varepsilon = \frac{1}{2} [\nabla \bf u + (\nabla \bf u)^{\it T} ],
\end{equation}

\noindent %where {\bf u} =  [$u,v$]\textsuperscript{\it T} is the displacement vector, 
\ReviewerC{where {\bf u} = $[u,v]^T$ is the displacement vector.} It can also be re-written as:

\begin{equation}
\varepsilon = \begin{bmatrix}
      		\varepsilon_{xx} = \frac{\partial u}{\partial x} & \varepsilon_{xy} = \frac{1}{2}(\frac{\partial u}{\partial y} + \frac{\partial v}{\partial x}) \\[1em]
      	    \varepsilon_{yx} = \frac{1}{2}(\frac{\partial v}{\partial x} + \frac{\partial u}{\partial y}) & \varepsilon_{yy} = \frac{\partial v}{\partial y}
     		\end{bmatrix},
\end{equation}

\noindent
where the diagonal strain components, ($\varepsilon_{xx},\varepsilon_{yy}$), are normal strain components and ($\varepsilon_{xy},\varepsilon_{yx}$) are shear strain components. \WKS{Specifically}, normal strain measures the changes in length along a specific direction, whereas shear strains measure \WKS{the} changes in two \christy{angular}.
The optical strain magnitude for each pixel can be calculated by taking the sum of squares of the normal and shear strain components, expressed below:

\begin{equation}
\begin{split}
|\varepsilon_{x,y}| &= \sqrt{{\varepsilon_{xx}}^{2} + {\varepsilon_{yy}}^{2} + {\varepsilon_{xy}}^{2} +{\varepsilon_{yx}}^{2}} \\
& = \sqrt{{\frac{\partial u}{\partial x}}^{2} + {\frac{\partial v}{\partial y}}^{2} +\frac{1}{2}{(\frac{\partial u}{\partial x} + \frac{\partial u}{\partial x})}^{2}}.
\end{split}
\label{eq:osm}
\end{equation}

\subsubsection{Bi-Weighted Oriented Optical Flow}
In this stage, we utilize the three aforementioned characteristics (i.e., orientation, magnitude and optical strain images for every video) to build a block-based Bi-Weighted Oriented Optical Flow. 

The three characteristic images are  partitioned equally into $N \times N$ non-overlapping blocks. For each block, the orientations $\theta_{x,y}\WKS{\in} [-\pi, \pi]$ are binned and locally weighted according to its magnitude $\rho_{x,y}$. Thus, the range of each histogram bin is:
\begin{equation}\label{eq:bin}
-\pi + \frac{2\pi c}{C} \leq \theta_{x,y} < -\pi + \frac{2\pi (c+1)}{C},
\end{equation}

\noindent where bin \WKS{$c \in \{1,2, \dots,C\}$}, and $C$ denotes the total number of histogram bins.

To obtain the global weight $\zeta_{b_{1},b_{2}}$ for each block, we utilize the optical strain magnitude $\varepsilon_{x,y}$ as follows:
\begin{equation}
\zeta_{b_{1},b_{2}} = \frac{1}{HL}\sum\limits_{y=(b_{2}-1)H+1}^{b_{2}H}\; \sum\limits_{x=(b_{1}-1)L+1}^{b_{1}L} \varepsilon_{x,y},
\end{equation}

\noindent where $L=\frac{X}{N}$, $H=\frac{Y}{N}$, the \ct{$b_1$ and $b_2$ are the block indices such that $b_{1},b_{2}\in 1,2, \ldots ,N$,} %\WKS{This expression does not make sense. Do you mean $b_1$, $b_2$ each assumes the values of $1, 2, ... N$? If that's the case, it should be expressed as \WKS{$b_1, b_2 \in \{1,2, \ldots N\}$}}
$X \times Y$ is the dimensions (viz., width-by-height) of the video frame. 

Lastly, the coefficients of $\zeta_{b_{1},b_{2}}$ are multiplied with the locally weighted histogram bins to their corresponding blocks. The histogram bins of each block are concatenated to form the resultant feature histogram.

In contrast to the conventional Histogram of Oriented Optical Flow (HOOF) \cite{chaudhry2009histogram}, \WKS{our proposed} orientation histogram bins have equal votes. Here, we consider both the magnitude and optical strain values as the weighting schemes to highlight the importance of each optical flow. Hence, a larger intensity of the pixel's movement or deformation contributes more effect \ct{to} the histogram, whereas noisy optical flows with small intensities \christy{reduce} %\WKS{It just has smaller contribution right? I think suppress is inappropriate. We suppress the noise, bit-stream size increment, overhead, usually negative stuffs.} 
 the significance of the features.

\begin{figure*}[t!]
\centering
\includegraphics[width=0.8\linewidth]{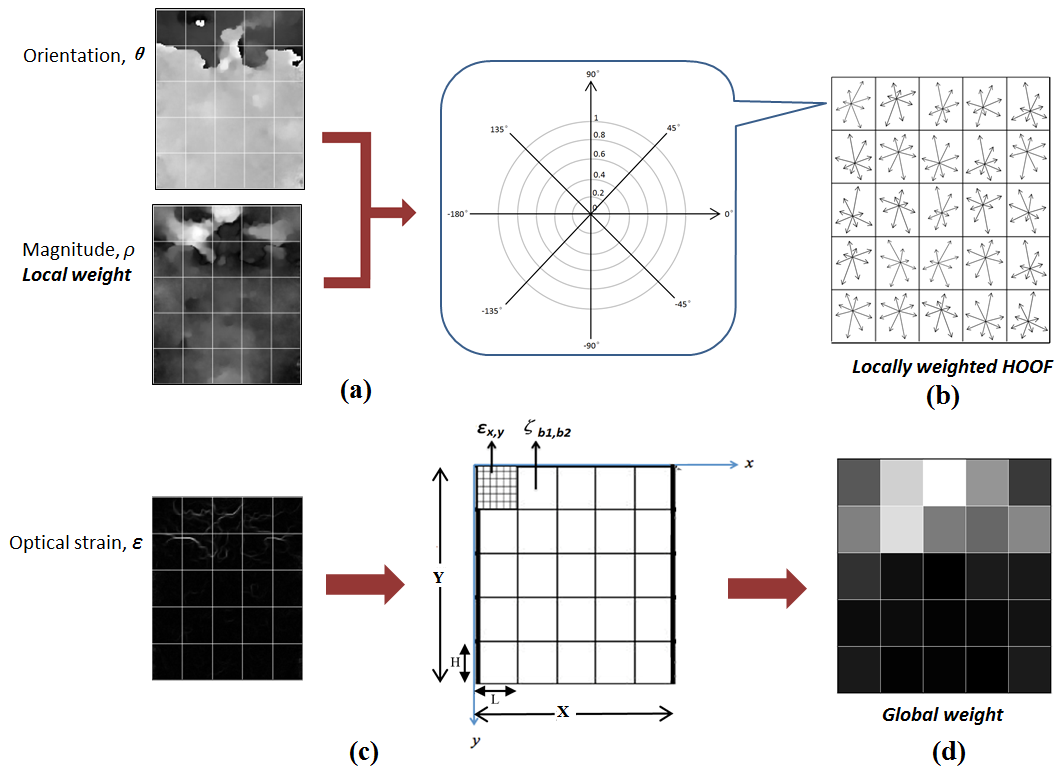}
\caption{The process of Bi-WOOF feature extraction for a video sample: (a) $\theta$ and $\rho$ images are divided into $N\times N$ blocks. In each block, the values of $\rho$ for each pixel are treated as local weights to multiply with their respective $\theta$ histogram bins; (b) It forms a locally weighted HOOF with feature size of $N\times N \times C$; (c) $\zeta_{b1,b2}$ denotes the global weighting matrix, which is derived from $\varepsilon$ image; (d) Finally, $\zeta_{b1,b2}$ are multiplied with their corresponding locally weighted HOOF.}
\label{fig:flow2}
\end{figure*}

The overall process flow of obtaining the locally and globally weighted features is illustrated in Figure~\ref{fig:flow2}.

\section{Experiment}
\label{sec:experiment}
\subsection{Datasets}
To evaluate the performance of the proposed algorithm, the experiments were carried out on \modified{five} \WKS{recent} spontaneous micro-expression databases, namely \modified{CAS(ME$)^2$}~\cite{qu2017casme}, CASME II \cite{casme2}, SMIC-HS \cite{smic}, SMIC-VIS \cite{smic} and SMIC-NIR \cite{smic}. Note that all these databases are recorded in a constrained laboratory condition due to the subtlety of micro-expressions.

\subsubsection{CASME II}
CASME II consists of five \WKS{classes} of expressions: surprise (25 samples), repression (27 samples), happiness (23 samples), disgust (63 samples) and others (99 samples). Each video clip contains only one micro-expression. Thus, there is a total of 246 video sequences. The emotion labels were marked by two coders with the reliability of 0.85. The expressions were elicited from 26 subjects with the mean age of 22 years old, \WKS{and recorded using the camera - Point Grey GRAS-03K2C}. The video resolution and frame rate of the camera are $640 \times 480$ pixels and 200 fps respectively. This database provides the cropped video sequences, \WKS{where only the face region is shown while} the unnecessary background has been eliminated. The cropped images have an average spatial resolution of $170 \times 140$ pixels, and each video consists of 68 frames (viz., $0.34s$). The video with the highest and lowest number of frames are 141 (viz., $0.71s$) and 24 (viz., $0.12s$), respectively. The frame index (i.e., frame number) for onset, apex and offset of each video sequence are provided. To perform the recognition task on this micro-expression dataset, the block-based LBP-TOP feature was considered. The features were then classified by a Support Vector Machine (SVM) with leave-one-video-out cross-validation (LOVOCV) protocol.

\subsubsection{SMIC}
SMIC includes three sub-datasets, which are SMIC-HS, SMIC-VIS and SMIC-NIR. The data composition of these datasets are detailed in Table~\ref{table:smic_datasets} %\WKS{This is not our database right? Can we push the table to Appendix?}\christy{we used this database for experiment too}
. It is noteworthy that all eight participants %\WKS{actors?}\christy{smic paper called it as ``participants"}
 who appeared in \WKS{the} VIS and NIR datasets were also involved in HS dataset elicitation. During the recording process, three cameras (i.e., HS, VIS and NIR) were recording simultaneously. The cameras were placed parallel to each other at the middle-top of the monitor. The groundtruth of the frame indices of onset and offset for each video clip in SMIC are given, but not the apex frame. The three-class recognition task was carried out for the three SMIC datasets individually %\WKS{What do you want to say here? three-class recognition refers to??}\christy{3-class refers to positive, negative and surprise classes. Each of the HR,VIS and NIR databases performed the recognition task} 
by utilizing block-based LBP-TOP as the feature extractor and SVM-LOSOCV (leave-one-subject-out cross-validation) as the classifier.

\subsubsection{CAS(ME$)^2$}
\modified{CAS(ME$)^2$ dataset has two major parts (A and B). 
Part A consists of 87 long videos, containing both spontaneous macro-expressions and micro-expressions.
Part B contains 300 short videos (i.e., cropped faces) spontaneous macro-expression samples and 57 micro-expression samples.
To evaluate the proposed method, we only consider the cropped micro-expression videos (i.e., 57 samples in total). 
However, we discovered three samples are missing from the dataset provided. 
Hence, 54 micro-expression video clips are used in the experiment.
The micro-expression video sequences are elicited from 14 participants. 
This dataset provides the cropped face video sequence. 
The videos are recorded using Logitech Pro C920 camera with a temporal resolution of 30 fps and spatial resolution of 640 $\times$ 480 pixels.
It composes of four classes of expressions: negative (21 samples), others (19 samples), surprise (8 samples) and positive (6 samples).
We resized the images to 170 $\times$ 140 pixels for experiment purpose.
The average number of frames of the micro-expression video sequences is 6 frames (viz., $0.2s$). The video with the highest and lowest number of frames are 10 (viz., $0.33s$) and 4 (viz., $0.13s$), respectively. 
The ground-truth frame indices for onset, apex and offset of each video sequence are also provided. 
To annotate the emotion label for each video sequence, a combination of AUs, emotion types of expression-elicitation video and self-reported are considered.
%The spotting rate of micro-expressions is around 40\% and the features are captured using LBP feature extractor.
The highest accuracy for the four-class recognition task reported in the original paper~\cite{qu2017casme} is 40.95\%.
It is obtained by adopting LBP-TOP feature extractor and SVM-LOSOCV classifier.}

\setlength{\tabcolsep}{4pt}
\begin{table*}[tb]
\begin{center}
\caption{Detailed information of the SMIC-HS, SMIC-VIS and SMIC-HR datasets}
\label{table:smic_datasets}
\begin{tabularx}{0.65\linewidth}{lXccc}
\hline
\noalign{\smallskip}
%\cline{3-5}
Datasets & & SMIC-HS & SMIC-VIS & SMIC-NIR \\
\noalign{\smallskip}
\hline
\noalign{\smallskip}

\multicolumn{2}{l} {Participants} 
&	16	&	8	&	8\\
\hline
\multirow{2}{*}{Camera} 
&	Type
&	\begin{tabular}{@{}c@{}}PixeLINK \\ PL-B774U\end{tabular}
&	\begin{tabular}{@{}c@{}}Visual \\ camera \end{tabular}
&	\begin{tabular}{@{}c@{}}Near-infrared \\ camera \end{tabular}\\

& Frame rate (fps)
&	100	&	25	& 25 \\
\hline
\multirow{4}{*}{Expression} 
&	Positive
  &	51	&	28	&	28 \\

& Negative
&	70	&	23	& 23 \\

& Surprise
&	43	&	20	& 20 \\

& Total
&	164	&	71	& 71 \\

\hline
%\multirow{2}{*}{Image resolution} 
\multirow{2}{*}{\begin{tabular}{@{}c@{}}Image \\ resolution\end{tabular}} 
 
&	Raw
&	$640 \times 480$
&	$640 \times 480$
&	$640 \times 480$ \\

& Cropped (avg.)
&	$170 \times 140$
&	$170 \times 140$
&	$170 \times 140$\\

\hline
\multirow{3}{*}{\begin{tabular}{@{}c@{}}Frame \\ number\end{tabular} } 
&	Average
&	34
&	10
&	10 \\

& Maximum
&	58
&	13
&	13 \\

& Minimum
&	11
&	4
&	4 \\

\hline
\multirow{3}{*}{\begin{tabular}{@{}c@{}}Video \\ duration ($s$)\end{tabular}} 
&	Average
&	0.34
&	0.4
&	0.4 \\

& Maximum
&	0.58
&	0.52
&	0.52 \\

& Minimum
&	0.11
&	0.16
&	0.16 \\
\hline
\end{tabularx}
\end{center}
\end{table*}
\setlength{\tabcolsep}{1.4pt}
\setlength{\tabcolsep}{4pt}

\subsubsection{Experiment Settings}
%In this section, we present the details of the measurement protocol of the experiments, specifically the evaluation method and the parameter settings in the proposed method.

The aforementioned databases (i.e., \modified{CAS(ME$)^2$}, CASME II and SMIC) have imbalance distribution of the emotion types. Therefore, it is necessary to measure the recognition performance of the proposed method using F-measure, \WKS{which was }also suggested in \cite{le2014spontaneous}. Specifically, F-measure is defined as:
\ReviewerA{
\begin{equation}\label{eq:f-measure}
\text{F-measure} := 2 \times \frac{\text{Precision} \times \text{Recall}}{\text{Precision + Recall}} 
\end{equation}
for
\begin{equation}\label{eq:recall}
\text{Recall} := \frac{\sum_{i=1}^M \text{TP}_i}{\sum_{i=1}^M\text{TP}_i + \sum_{i=1}^M \text{FN}_i} 
\end{equation}
and
\begin{equation}\label{eq:precision}
\text{Precision} := \frac{\sum_{i=1}^M \text{TP}_i}{\sum_{i=1}^M\text{TP}_i + \sum_{i=1}^M \text{FP}_i} 
\end{equation}}

\noindent 
where $M$ is the number of classes; TP, FN and FP are the true positive, false negative and false positive, respectively.

On the other hand, to avoid person dependent issue in the classification process, we employed LOSOCV strategy in the linear SVM classifier setting. In LOSOCV, \christy{the features of the sample videos in one subject are treated as the testing data and the remaining features from rest of the subjects \WKS{become} the training data}%\WKS{Tak faham... Please explain / rephrase.}
. Then, this process is repeated for $k$ times, where $k$ is the number of subjects in the database.  Finally, the recognition results for all the subjects are averaged to compute the recognition rate.

For the block-based feature extraction methods (i.e., LBP, LBP-TOP and proposed algorithm), we standardized the block sizes to $5\times 5$ and $8\times 8$ for the SMIC and CASME II datasets, respectively, as we discovered that these block settings generated reasonably good recognition performance in all cases.
\modified{Since CAS(ME$)^2$ was only made public recently, there is still no method designed and tested on this dataset in the literature.
Hence, we report the recognition results for various block sizes using the baseline LBP-TOP and our proposed Bi-WOOF methods.}
\section{Results and Discussion}  
\label{sec:results}

In this section, we present the recognition results with detailed analysis and benchmarking against state-of-the-art methods. We also examine the computational efficiency of our proposed method, and lay down some key propositions derived from observations in this work.

\subsection{Recognition Results}

\modified{We report the results in two parts, according to the databases: (i)  CAS(ME$)^2$ (in Table~\ref{table:casmesquare}) and (ii) CASME II, SMIC-HS, SMIC-VIS and SMIC-NIR (in Table~\ref{table:results})}.

\modified{Table~\ref{table:casmesquare} records the recognition performance on CAS(ME$)^2$ with various block sizes by employing the baseline LBP-TOP and our proposed Bi-WOOF feature extractors.
This is because the original paper~\cite{qu2017casme} did not perform recognition task solely on the micro-expression samples, instead the result reported was tested on the mixed macro-expression and micro-expression samples.
We record both the F-measure and Accuracy measurements for different blocks sizes, including 5 $\times$ 5, 6 $\times$ 6, 7 $\times$ 7 and 8 $\times$ 8 for both feature extraction methods.
The best F-measure performance achieved by LBP-TOP is 41\%, while Bi-WOOF method achieves 47\%. 
Both results are obtained when block size is set to 6 $\times$ 6.
}

\setlength{\tabcolsep}{4pt}
\begin{table}[tb!]
\begin{center}
\caption{Micro-expression recognition results (\%) on CAS(ME$)^2$ with different number of block size for the LBP-TOP and Bi-WOOF feature extractors
\label{table:casmesquare}}
\begin{tabular}{lcccc}
\noalign{\smallskip}
\cline{2-5} 
& \multicolumn{2}{c}{F-measure} & \multicolumn{2}{c}{Accuracy} \\
\cline{1-5} 
Block & LBP-TOP & Bi-WOOF & LBP-TOP & Bi-WOOF  \\
\hline
\noalign{\smallskip}

5x5
& .28	& \textbf{.47} & 46.30	& \textbf{59.26}	\\

6x6
& .41	& \textbf{.47} & 48.15 & \textbf{59.26}	\\

7x7
& .26	& \textbf{.46} & 44.44 & \textbf{59.26}	\\

8x8
& .28	& \textbf{.47} & 48.15 & \textbf{59.26}	\\

\hline
\end{tabular}
\end{center}
\end{table}
\setlength{\tabcolsep}{1.4pt}
\setlength{\tabcolsep}{4pt}

\modified{The micro-expression recognition performances of the proposed method (i.e., Bi-WOOF) and the other conventional feature extraction methods evaluated on CASME II, SMIC-HS, SMIC-VIS and SMIC-NIR databases are shown in Table~\ref{table:results}.}
Note that the \emph{sequence-based} methods \#1 to \#13 considered all \WKS{frames} in the video sequence (i.e., frames from onset to offset). Meanwhile, methods \#14 to \#19 consider only information from the apex and onset frames, whereby only two images are processed to extract features. We refer to these as \emph{apex-based} methods. 

% js: describe the need for spotting the apex
%The first and most essential criteria to employ the 
\john{Essentially, our proposed apex-based approach requires determining} the apex \john{frame} %index (i.e., frame number) 
for each video sequence. Although the SMIC datasets (i.e., HS, VIS and NIR) did not provide the ground-truth apex frame indices, we utilize the \emph{divide-and-conquer} strategy proposed in~\cite{liong2015automatic} to \john{spot} the apex frame. \john{For CASME II, the ground-truth apex frame indices are already provided, so we can use them directly.} %As such, the lack of the apex frame information issue had been resolved. 

In order to \john{validate} the importance of the apex frame, we also randomly select one frame \john{from} each video sequence. \john{Features are then computed using the apex/ random frame and the onset (reference) frame} using LBP \christyy{, HOOF} and Bi-WOOF descriptors. The recognition performances of the random frame selection approaches (repeated for 10 times) are \WKS{reported as methods \#14\christyy{, \#16} and \#18} \john{while the apex-frame approaches are reported as methods \#15, \#17 and \#19.}  %This process was repeated for 10 times. 
We observe that the utilization of the apex frame always yields better \WKS{recognition results} when compared to using random frames. As such, it can be concluded that the apex frame plays an important role in \WKS{forming discriminative} features.

For method \#1 (i.e., LBP-TOP), also \ct{referred to as} the baseline, we reproduced the experiments for the four datasets based on the original papers~\cite{smic,casme2}. The recognition rates for methods \#2 to \#11 are reported from their respective works of the same experimental protocol. 
\modified{Besides, we replicated method \#12 and evaluate it on CASME II database. This is because the original paper~\cite{liu2016main} classifies the emotion into 4 types (i.e., positive, negative, surprise and others). 
For a fair comparison with our proposed method, we re-categorize the emotions into 5 types (i.e., happiness, disgust, repression, surprise and others).}
\john{For method \#13,} Bi-WOOF is applied on all frames in the video sequence. The features were computed by first estimating the three characteristics of the optical flow (i.e., orientation, magnitude and strain) between the onset and each subsequent frame (i.e., $\{f_{i,1}, f_{i,j}\}, j \in 2,\dots,F_i$ ). Next, Bi-WOOF was \john{computed} for each \john{pair of frames} to obtain the resultant histogram. %The recognition performance is reported under method \#12.

%For the LBP feature extractor (i.e., methods \#13 and \#14), 
\modifiedA{LBP was applied on the difference image to compute the features in methods \#14 and \#15.
Note that the image subtraction process is only applicable for methods \#14 (LBP - random \& onset) and \#15 (LBP - apex \& onset).
This is because LBP feature extractor can only capture the spatial features of an image and it is incapable of extracting the temporal features of two images.
Specifically, the spatial features extracted from the apex frame and the onset frame are not correlated. 
Hence, we perform an image subtraction process in order to generate a single image from two images (i.e., apex / random frame and onset frame).
This image subtraction process can remove a person's identity while preserving the characteristics of facial micro-movements.
Besides, for the apex-based approaches, we also evaluated the HOOF feature (i.e., methods \#16 and \#17) by binning the optical flow orientation, which is computed between the apex / random frame and the onset frame, to form the feature histogram.}

Table \ref{table:results} suggests that the proposed algorithm (viz., \#19) achieves promising results in all four datasets. More \ct{precisely}, it outperformed all the other methods in CASME II. In addition, for SMIC-VIS and SMIC-NIR, the results of the proposed method are comparable to those of \#9, viz., FDM method.

\setlength{\tabcolsep}{4pt}
\begin{table*}[tb!]
\begin{center}
\caption{Comparison of micro-expression recognition performance in terms of F-measure on the CASME II, SMIC-HS, SMIC-VIS and SMIC-NIR databases for the state-of-the-art feature extraction methods, and the proposed apex frame methods}
\label{table:results}
\begin{tabular}{lllcccc}
\noalign{\smallskip}
\cline{1-7}
\noalign{\smallskip}
&& Methods & CASME II & SMIC-HS & SMIC-VIS & SMIC-NIR \\
\noalign{\smallskip}
\hline
\noalign{\smallskip}

\multirow{13}{*}{\rotatebox{90}{Sequence-based}}
%\multirow{10}{*}{\parbox{1.7cm}{Whole \\sequence \\ (i.e., multiple \\ frames)}}

&	1	&	LBP-TOP \cite{smic,casme2}	& .39	&	.39	& .39	& .40	\\
&	2	&	OSF \cite{liong2014optical}	& -	&	.45	& -	& -	\\
&	3	&	STM \cite{le2014spontaneous}	& .33	&	.47	& -	& -	\\
&	4	&	OSW \cite{liong2014subtle}	& .38	&	.54	& -	& -	\\
&	5	&	LBP-SIP	\cite{wang2015lbp} & .40	&	.55	& -	& -	\\
&	6	&	MRW \cite{oh2015monogenic}	& .43	&	.35	& -	& -	\\
&	7	&	STLBP-IP \cite{huang2015facial}	& .57	&	.58	& -	& -	\\
&	8	&	OSF+OSW \cite{liong2016spontaneous}	& .29	&	.53	& -	& -	\\
&	9	&	FDM \cite{xu2016microexpression}	& .30	&	.54	& .60	& .60	\\
&	10	&	\begin{tabular}{@{}l@{}}Sparse  \\ Sampling \cite{le2016sparsity} \end{tabular} 	& .51	&	.60	& -	& -	\\
&	11	&	STCLQP \cite{huang2016spontaneous}	& .58	&	.64	& -	& -	\\
&	12	&	MDMO~\cite{liu2016main} & .44	&	-	& -	& -	\\
&	13	&	Bi-WOOF	& .56	&	.53	& .62	& .57	\\

\noalign{\smallskip}
\hline
\noalign{\smallskip}
%\multirow{4}{*}{\parbox{1.7cm}{2 images\\ (apex \& onset)} }
\multirow{12}{*}{\rotatebox{90}{Apex-based}}
&	14	&	\begin{tabular}{@{}l@{}}LBP \\(random \& onset) \end{tabular}  	& .38	&	.40	& .48	& .51	\\
&	15	&	\begin{tabular}{@{}l@{}}LBP \\ (apex \& onset) \end{tabular} 	& .41	&	.45	& .49	& .54\\
&	16	&	\begin{tabular}{@{}l@{}}HOOF \\ (random \& onset) \end{tabular} 	& .41	&	.40	& .51	& .50	\\
&	17	&	\begin{tabular}{@{}l@{}}HOOF \\ (apex \& onset) \end{tabular}	& .43	&	.48	& .49	& .47\\
&	18	&	\begin{tabular}{@{}l@{}}Bi-WOOF \\ (random \& onset) \end{tabular}	& .50	&	.46	& .56	& .50	\\
&	19	&\bf \begin{tabular}{@{}l@{}}Bi-WOOF \\ (apex \& onset) \end{tabular} & \bf .61	&\bf .62	&\bf .58	&\bf .58	\\
\hline
\end{tabular}
\end{center}
\end{table*}
\setlength{\tabcolsep}{1.4pt}
\setlength{\tabcolsep}{4pt}

\subsection{Analysis and Discussion}
%\subsection{Discussions}
%\subsubsection{Detailed Analysis on the Recognition Performance}

\modified{To further analyze the recognition performances, we provide the confusion matrices for the selected databases.
Firstly, for CAS(ME$)^2$, as tabulated in Table~\ref{table:mat_casmesquare}, it can be seen that the recognition rate using Bi-WOOF method outperforms LBP-TOP method for all block sizes. 
Therefore, it can be concluded that the Bi-WOOF method is superior compared to the baseline method.}

\modified{On the other hand, for the CASME II and SMIC databases, we only present the confusion matrices for the high frame rate databases, namely, CASME II and SMIC-HS. 
This is because most works in the literature tested on these two spontaneous micro-expression databases, making performance comparisons possible.}
\modifiedB{It is worth highlighting that a number of works in literature such as~\cite{wang2015micro,liu2016main}, perform classification of micro-expressions in CASME II based on four categories (i.e., negative, positive, surprise and others), instead of the usual five (i.e., disgust, happiness, tense, surprise and repression) as used in most works.}

\modified{The confusion matrices are recorded in Tables~\ref{table:mat_casme} and \ref{table:mat_smic} for CASME II and SMIC-HS, respectively.} 
It is observed that there are significant improvements in  classification performance for all kinds of expression when employing Bi-WOOF (apex \& onset) when compared to the baselines. More concretely, in CASME II, the recognition rate of surprise, disgust, repression, happiness and other expressions were \ct{improved} by 44\%, 30\%, 22\%, 13\% and 4\%, respectively. Furthermore, for SMIC-HS, the recognition rate of the expressions of negative, surprise and positive were improved by 31\%, 19\% and 18\%, respectively.

%%%%%%%%%%%%% c_t - CASME^2%%%%%%%%%%%%%%%%%%%
\begin{table}[tb!]
	\begin{center}
    \caption{Confusion matrices of baseline and Bi-WOOF (apex \& onset) for the recognition task on CAS(ME$^2)$ database for block size of 6, where the emotion types are, POS: positive; NEG: negative; SUR: surprise; OTH: others}
    \label{table:mat_casmesquare}
    \begin{subtable}{1\linewidth}\centering
    %\begin{subtable}{\textwidth}\centering
      %\centering
        \caption{Baseline}     
        \begin{tabular}{lcccc}
        \noalign{\smallskip}
        \cline{2-5} 
         & POS	 & NEG	&	SUR	&	OTH	\\
        \noalign{\smallskip}
        \hline
        \noalign{\smallskip}

        POS  
        &\textbf{.17}	&	.33	&	0	&	.50	\\

        NEG  
        & 0	&	\textbf{.67}	&	0	&	.33	\\
        
        SUR
        & 0	&	.38	&	\textbf{0}	&	.63	\\
        
        OTH  
        & 0	&	.42	&	0	&	\textbf{.58}	\\
        
        \hline
        \end{tabular}
    \end{subtable}%    
    \vskip 8pt
    \begin{subtable}{1\linewidth}\centering
    %\begin{subtable}{\textwidth}\centering
      %\centering
        \caption{Bi-WOOF (apex \& onset)}
        \begin{tabular}{lcccc}
        \noalign{\smallskip}
        \cline{2-5} 
         & POS	 & NEG	&	SUR	&	OTH	\\
        \noalign{\smallskip}
        \hline
        \noalign{\smallskip}

        POS  
        &\textbf{0}	&	0	&	.50	&	.50	\\

        NEG  
        & 0	&	\textbf{.71}	&	.05	&	.24	\\
        
        SUR
        & .25	&	.13	&	\textbf{.50}	&	.13	\\
        
        OTH  
        & .16	&	.16	&	0	&	\textbf{.68}	\\
        
        \hline
        \end{tabular}
    \end{subtable} 
    \end{center}
\end{table}

%%%%%%%%%%%%% c_t - CASME II %%%%%%%%%%%%%%%%%%%
\begin{table}[tb!]
	\begin{center}
    \caption{Confusion matrices of baseline and Bi-WOOF (apex \& onset) for the recognition task on CASME II database, where the emotion types are, DIS: disgust; HAP: happiness; OTH: others; SUR: surprise; and REP:repression}
    \label{table:mat_casme}
    \begin{subtable}{1\linewidth}\centering
    %\begin{subtable}{\textwidth}\centering
      %\centering
        \caption{Baseline}     
        \begin{tabular}{lccccc}
        \noalign{\smallskip}
        \cline{2-6} 
         & DIS	 & HAP	&	OTH	&	SUR	&	REP \\
        \noalign{\smallskip}
        \hline
        \noalign{\smallskip}

        DIS  
        &\bf.20	&	.11	&	.66	&	.02	&	.02\\

        HAP 
        &	.09	&\bf.47	&	.25	&	0	&	.19	\\
        
        OTH 
        &	.21	&	.12	&\bf.58	&	.08	&	0	\\
        
        SUR 
        &	.12	&	.36	&	.20	&\bf.32	&	0	\\
       
       	REP 
        &	.07	&	.33	&	.26	&	.04	&\bf.30	\\
        \hline
        \end{tabular}
    \end{subtable}%    
    \vskip 8pt
    \begin{subtable}{1\linewidth}\centering
    %\begin{subtable}{\textwidth}\centering
      %\centering
        \caption{Bi-WOOF (apex \& onset)}
        \begin{tabular}{lccccc}
        \noalign{\smallskip}
        \cline{2-6} 
        & DIS	 & HAP	&	OTH	&	SUR	&	REP \\
        \noalign{\smallskip}
        \hline
        \noalign{\smallskip}

        DIS  
        &\bf.49	&	.07	&	.44	&	0	&	0\\

        HAP 
        &	.03	&\bf.59	&	.28	&	.03	&	.06	\\
        
        OTH 
        &	.21	&	.09	&\bf.62	&	.01	&	.06	\\
        
        SUR 
        &	.04	&	.12	&	.08	&\bf.76	&	0	\\
       
       	REP 
        &	.07	&	.19	&	.22	&	0	&\bf.52	\\
        \hline
        \end{tabular}
    \end{subtable} 
    \end{center}
\end{table}

%%%%%%%%%%%%% c_t - SMIC  %%%%%%%%%%%%%%%%%%%

\begin{table}[tb!]
	\begin{center}
    \caption{Confusion matrices of baseline and Bi-WOOF (apex \& onset) for the recognition task on SMIC-HS database, where the emotion types are, NEG: negative; POS: positive; and SUR:surprise}
    \label{table:mat_smic}
    \begin{subtable}{1\linewidth}\centering
    %\begin{subtable}{\textwidth}\centering
      %\centering
        \caption{Baseline}     
        \begin{tabular}{lccc}
        \noalign{\smallskip}
        \cline{2-4} 
         & NEG	 & POS	&	SUR	\\
        \noalign{\smallskip}
        \hline
        \noalign{\smallskip}

        NEG  
        &\bf.34	 & .29	&	.37	\\

        POS 
        & .41	 &\bf.39	&	.20	\\
        
        SUR 
        & .37	 & .19	&\bf.44	\\
        
        \hline
        \end{tabular}
    \end{subtable}%    
    \vskip 8pt
    \begin{subtable}{1\linewidth}\centering
    %\begin{subtable}{\textwidth}\centering
      %\centering
        \caption{Bi-WOOF (apex \& onset)}
        \begin{tabular}{lccccc}
        \noalign{\smallskip}
        \cline{2-4} 
         & NEG	 & POS	&	SUR	\\
        \noalign{\smallskip}
        \hline
        \noalign{\smallskip}

        NEG  
        &\bf.66	 & .23	&	.11	\\

        POS 
        & .27	 &\bf.57	&	.16\\
        
        SUR 
        & .23	 & .14	&\bf.63	\\
        \hline
        \end{tabular}
    \end{subtable} 
    \end{center}
\end{table}

\begin{figure*}[tb!]
\centering
        \begin{subfigure}[b]{.19\textwidth}
                \includegraphics[width=\linewidth]{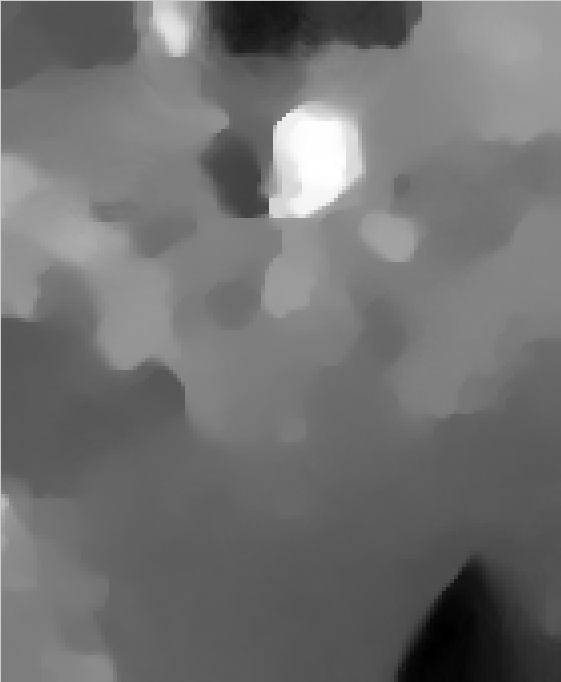}
                \caption{$p$}
                \label{fig:u}
        \end{subfigure}%
        \hspace{\fill}
        \begin{subfigure}[b]{.19\textwidth}
                \includegraphics[width=\linewidth]{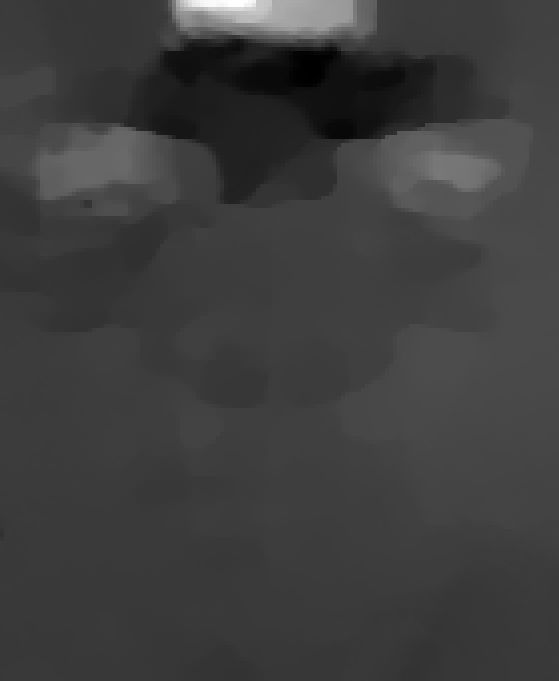}
                \caption{$q$}
                \label{fig:v}
        \end{subfigure}%
        \hspace{\fill}
        \begin{subfigure}[b]{.19\textwidth}
                \includegraphics[width=\linewidth]{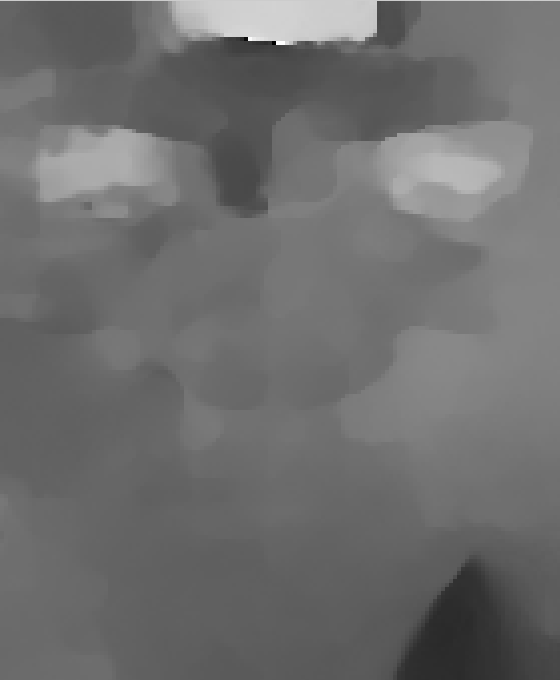}
                \caption{$\theta$}
                \label{fig:ori}
        \end{subfigure}
        \hspace{\fill}
        \begin{subfigure}[b]{.19\textwidth}	
                \includegraphics[width=\linewidth]{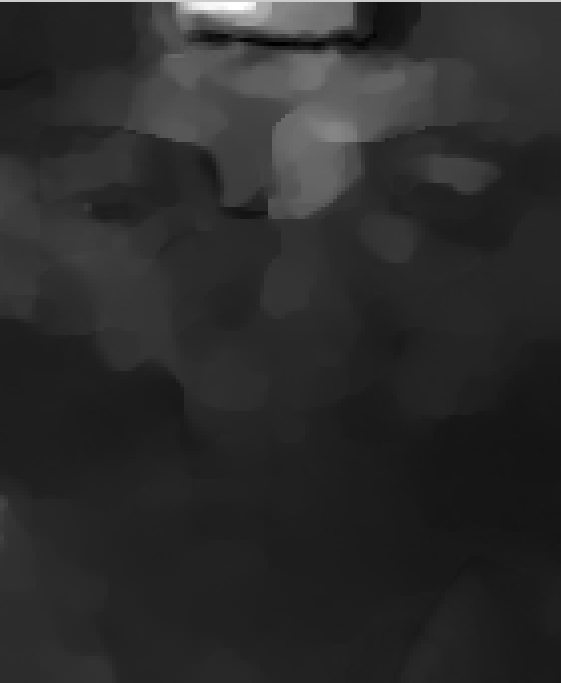}
                \caption{$\rho$}
                \label{fig:mag}
        \end{subfigure}%
        \hspace{\fill}
        \begin{subfigure}[b]{.19\textwidth}
                \includegraphics[width=\linewidth]{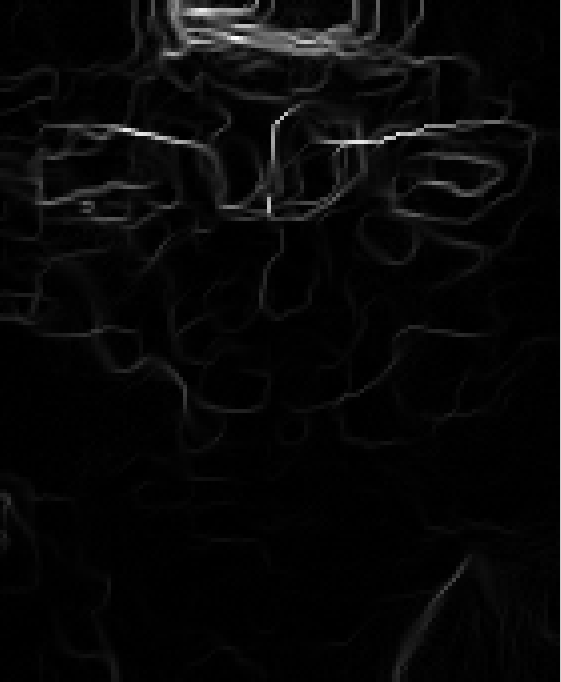}
                \caption{$\varepsilon$}
                \label{fig:os}
        \end{subfigure}
        \caption{Illustration of \WKS{components derived from optical flow using} onset and apex frames of a video: (a) Horizontal vector of optical flow, $p$; (b) Vertical vector of optical flow, $q$; (c) Orientation, $\theta$; (d) Magnitude, $\rho$; (e) Optical strain, $\varepsilon$}
\label{fig:illustration} % s04_sur_01
\end{figure*}

Figure \ref{fig:illustration} exemplifies the components derived from optical flow using onset and apex frames of the video \christy{sample ``s04{\_}sur{\_}01'' in SMIC-HS}%\WKS{XXX - provide specific video information}
, where the micro-expression of \emph{surprise} is shown. Referring to the labeling criteria of the emotion in~\cite{casme2}, the changes in facial muscles are centering at the eyebrow regions. We can hardly tell the facial movements in Figure~\ref{fig:u}, \ref{fig:v} and \ref{fig:ori}. For Figure~\ref{fig:mag}, a noticeable amount of the muscular changes are occurring at the upper part of the face, whereas in Figure~\ref{fig:os}, the eyebrows regions have obvious facial movement. Since magnitude information emphasizes the amplitude of the facial changes, we exploit it as local weight. \WKS{Due to the computation of higher order derivatives} in obtaining the optical strain magnitudes, \christy{optical strain }%\WKS{'it' refers to what?}
has the ability to remove the noise and preserve large motion changes. We exploit these characteristics to build the global weight. In addition, \cite{liong2014optical} demonstrated that optical strain globally weighted on the LBP-TOP features produced better recognition results when compared to results obtained without the weighting.

\WKS{Based on the results of }F-measure and confusion matrices, \WKS{it is observed} that extracting the features of two images only (i.e., apex and onset frame) using the proposed method (i.e., Bi-WOOF) is able to yield superior recognition performance for the micro-expression databases considered, especially in CASME II and SMIC-HS, which have high temporal resolution (i.e., $\geq$ 100 fps). 

\modifiedA{The number of histogram bins $C$ in Eq.~(\ref{eq:bin}) is empirically determined to be 8 for both the CASME II and SMIC-HS databases.
Table \ref{table:bins} quantitatively illustrates the relationship between the recognition performance and the histogram bins.
%Besides, Figure~\ref{fig:graph_f1} and Figure~\ref{fig:graph_acc} graphically show the F-measure and Accuracy performance with different number of histogram bins, respectively.
It can be seen that with histogram bin = 8, the Bi-WOOF feature extractor achieves the best recognition results on both CASME II and SMIC-HS databases.}

\modifiedA{We provide in Table \ref{table:weights} a closer look into the effects of applying (and not applying) the global and local weighting schemes on the Bi-WOOF features. Results on both SMIC-HS and CASME II are in agreement that the flow orientations are best weighted by their magnitudes, while the strain magnitudes are suitable as weights for the blocks. Results are the poorest when no global weighting is applied, which shows the importance of altering the prominence of features in different blocks.}

\setlength{\tabcolsep}{4pt}
\begin{table}[tb!]
\begin{center}
\caption{\ReviewerA{Micro-expression recognition results (\%) on SMIC-HS and CASME II databases with different number of histogram bins used for the Bi-WOOF feature extractor
\label{table:bins}}}
\begin{tabular}{lcccc}
\noalign{\smallskip}
\cline{2-5} 
& \multicolumn{2}{c}{CASME II} & \multicolumn{2}{c}{SMIC-HS} \\
\cline{1-5} 
Bin & F-measure & Accuracy & F-measure & Accuracy  \\
\hline
\noalign{\smallskip}

1 
& .39	& 46.09 & .46 & 45.12	\\

2 
& .61	& 57.20 & .50 & 50.00	\\

3
& .59	& 55.56 & .49 & 48.78	\\

4
& .54	& 51.03  & .58 & 58.54	\\

5 
& .60	& 58.02 & .53 & 54.27	\\

6 
& .58	& 54.32 & .54 & 54.27	\\

7
& .57	& 54.32 & .50 & 50.00	\\

8
& \bf{.61}	& \bf{58.85}	& \bf{.62} & \bf{62.20}\\

9
& .59	& 56.38	& .49 & 49.39\\

10
& .61	& 59.67	& .59 & 58.54\\

\hline
\end{tabular}
\end{center}
\end{table}
\setlength{\tabcolsep}{1.4pt}
\setlength{\tabcolsep}{4pt}

\begin{table}[tb!]
	\begin{center}
    \caption{Recognition performance (F-measure) with different combination of local and global weights used for Bi-WOOF}
    \label{table:weights}
    \begin{subtable}{1\linewidth}\centering
    %\begin{subtable}{\textwidth}\centering
      %\centering
        \caption{SMIC-HS}     
        \begin{tabular}{ccccc}
        \noalign{\smallskip}
        \cline{3-5}
        \noalign{\smallskip}
        & & \multicolumn{3}{c}{Local} \\
        \noalign{\smallskip}
        & Weights & None & Flow & Strain \\
        \noalign{\smallskip}
        \hline
        \noalign{\smallskip}

\multirow{3}{*}{\rotatebox{90}{Global}}

&	None	&	.44	& .42	&	.43 \\
&	Flow	&	.51 & .52	&	.50 \\
&	Strain	&	.54 & \bf{.62}	&	.59 \\
\noalign{\smallskip}
\hline
\noalign{\smallskip}
\end{tabular}
    \end{subtable}%    
    \vskip 8pt
    \begin{subtable}{1\linewidth}\centering
    %\begin{subtable}{\textwidth}\centering
      %\centering
        \caption{CASME II}
 		\begin{tabular}{ccccc}
      \noalign{\smallskip}
      \cline{3-5}
      \noalign{\smallskip}
      & & \multicolumn{3}{c}{Local} \\
      \noalign{\smallskip}
      & Weights & None & Flow & Strain \\
      \noalign{\smallskip}
      \hline
      \noalign{\smallskip}

      \multirow{3}{*}{\rotatebox{90}{Global}}

      &	None	&	.43	& .52	&	.49 \\
      &	Flow	&	.53 & .58	&	.56 \\
      &	Strain	&	.59 & \bf{.61}	&	.59 \\
      \noalign{\smallskip}
      \hline
      \noalign{\smallskip}
      \end{tabular}	
    \end{subtable} 
    \end{center}
\end{table}

\subsection{Computational Time}
We examine the computational efficiency of Bi-WOOF in SMIC-HS database on both the {\it whole sequence} and {\it two images (i.e., apex and onset)}, which are the methods \#1 and \#15 in Table~\ref{table:results}, respectively. The average duration taken per video \ct{for the execution of the} micro-expression recognition system for the {\it whole sequence} and {\it two images} in MATLAB implementation were 128.7134$s$ and 3.9499$s$ respectively. The time considered for this recognition system includes: (1) \ct{Spotting the apex frame} using the \emph{divide-and-conquer} strategy; (2) \ct{Estimation of the horizontal} and vertical components of optical flow; (3) \ct{Computation of} orientation, magnitude and optical strain images; (4) \ct{Generation of} Bi-WOOF histogram; (5) Expression classification in SVM. Both experiments were carried out on an Intel Core i7-4770 CPU 3.40GHz processor. Results suggest that the case of {\it two images} is $\sim$33 times faster than the case of {\it whole sequence}. It is indisputable that extracting the features from only \textit{two images} is significantly faster than the \textit{whole sequence} because lesser images are involved in the computation, and hence the volume of data to process is less.

\subsection{\it ``Prima facie"}
%john: decided to move this to the subsection of section 5 after discussions

At this juncture, we have established two strong propositions, which are by no means conclusive as further extensive research can provide further validation: 
%\WKS{WKS: Guess we can directly replace the current conclusion section by your texts / paragraphs}
\begin{enumerate}
	\item {\bf The apex frame is the most important frame in a micro-expression clip}, that it contains the most intense or expressive micro-expression information. \john{Ekman's \cite{ekman1993facial} and Esposito's \cite{esposito2007amount} suggestions are validated by our use of the apex frame to characterize the change in facial contraction, a property best captured by the proposed Bi-WOOF descriptor which considers both facial flow and strain information.} \john{Control} experiments \ct{using} random frame selection (as the supposed apex frame) substantiates this fact. Perhaps, in future work, it will be interesting to know to what extent an imprecise apex frame (for instance, a detected apex frame that is located a few frames away) could influence the recognition performance. Also, further insights into locating the apices of specific facial Action Units (AUs) could possibly provide even better discrimination between types of micro-expressions. 
	
    \item {\bf The apex frame is sufficient for micro-expression recognition.} A majority of recent state-of-the-art methods promote the use of the entire video sequence, or a reduced set of frames \cite{smic,le2016sparsity}. In this work, we advocate the opposite idea that, ``less is more'', supported by our hypothesis that a large number of frames does not guarantee a high recognition accuracy, particularly in the case when high-speed cameras are employed (e.g., for CASME II and SMIC-HS datasets). Comparisons against conventional sequence-based methods show that the use of the apex frame can provide more valuable information than a series of frames, what more at a much lower cost. At this juncture, it is premature to ascertain specific reasons behind this finding. Future directions point towards a detailed investigation into \emph{how} and \emph{where} micro-expression cues reside within the sequence itself. 
% js: I think it will do this paper a favour by adding apex spotting results on the onset-offset sequences. Some overlap with ACPR is ok, but it provides completeness to this paper.  
%Next, it is also questionable whether the reliability or the accuracy of labeling of the apex frame index (either automatically or manually) can be justified by the recognition performance.
\end{enumerate}
\section{Conclusion} 
\label{sec:conclusion}
In the recent few years, a number of research groups have attempted to improve the accuracy of micro-expression recognition by designing a variety of feature extractors that can best capture the subtle facial changes \cite{wang2015lbp,huang2015facial,liu2016main}, while a few other works \cite{le2015subtle,le2016sparsity,smic} have sought out ways to reduce information redundancy in micro-expressions (using only a portion of all frames) before recognizing them. %Most of them utilized whole video sequences while some used part of the frames of the video sequence in the feature extraction process. 

In this paper, we demonstrated that it is sufficient to encode facial micro-expression features by utilizing only the apex frame (and onset frame as reference frame). To the best of our knowledge, this is the first attempt at recognizing micro-expressions \ct{in} video using only the apex frame. For databases that do not provide apex frame annotations, the apex frame can be acquired by automatic spotting \ct{method} based on a \emph{divide-and-conquer} search strategy \WKS{proposed} in our recent work \cite{liong2015automatic}. We also proposed a novel feature extractor, namely, Bi-Weighted Oriented Optical Flow (Bi-WOOF), which can concisely describe discriminately weighted motion features extracted from the apex and onset frames. As its name implies, the optical flow histogram features (bins) are locally weighted by their own magnitudes while facial regions (blocks) are globally weighted by the magnitude of optical strain \christy{--} %\WKS{which is?}\christy{OS}
 a reliable measure of subtle deformation. 

Experiments conducted on \modified{five} publicly available micro-expression databases, namely, \modified{CAS(ME$)^2$}, CASME II, SMIC-HS, SMIC-NIR and SMIC-VIS, demonstrated the effectiveness and efficiency of the proposed approach. Using a single apex frame for micro-expression recognition, the two high frame rate databases, i.e., CASME II and SMIC-HS, both achieved the promising recognition rate of $61\%$ and $62\%$, respectively, when compared to the state-of-the-art methods.

\section*{Acknowledgments}
The authors would like to thank colleagues and anonymous reviewers for their valuable comments and suggestions to improve the quality of the paper. This work was supported in part by Telekom Malaysia R\&D Grant under Project 2beAware ((MMUE/140098), MOHE Grant FRGS/1/2016/ICT02/MMU/02/2, Multimedia University and University of Malaya.

\section*{References}
\bibliography{mybibfile}

\end{document}